\documentclass{article}
\usepackage{spconf,amsmath,graphicx,hyperref}
\usepackage{array} 

\usepackage{booktabs}
\usepackage{tabularx}
\usepackage{array,multirow}
\usepackage{enumitem}

\usepackage{threeparttable}

\title{SubQRAG: Sub-question driven Dynamic Graph RAG}
\name{
\parbox{\linewidth}{\centering
Jiaoyang Li$^{1,*}$\thanks{$*$ Equal contribution. $\dagger$ Corresponding author.}, 
Junhao Ruan$^{1,*}$, 
Shengwei Tang$^{1}$, 
Saihan Chen$^{1}$, 
Kaiyan Chang$^{1}$, 
Yuan Ge$^{1}$, \\
\textit{Tong Xiao}$^{1,2,\dagger}$, 
\textit{Jingbo Zhu}$^{1,2}$
}}

\address{$^{1}$ Northeastern University, Shenyang, China \\
         $^{2}$ NiuTrans Research, Shenyang, China}

\begin{document}
\maketitle
%

\begin{abstract}
Graph Retrieval-Augmented Generation (Graph RAG) is effective at building a knowledge graph (KG) to connect disparate facts across a wide document corpus.
However, this broad-view approach often lacks the deep structured reasoning required for complex multi-hop question answering (QA), leading to incomplete evidence and error accumulation. To address these issues, we propose SubQRAG\footnotemark, a sub-question driven framework that enhances reasoning depth. SubQRAG decomposes a complex question into an ordered chain of verifiable sub-questions. For each sub-question, it retrieves relevant triples from the graph. A key feature is its ability to adapt: if the original graph is insufficient, the system retrieves information from source documents, extracts new triples, and dynamically updates the graph in real time. Finally, all triples used in the reasoning process are aggregated into a ``graph memory'', providing a structured and traceable evidence path for the final answer generation. Experiments on three multi-hop QA benchmarks show that SubQRAG consistently achieves significant improvements, particularly in Exact Match scores.
\end{abstract}

\section{Introduction}
\label{sec:intro}

Large Language Models (LLMs) have demonstrated remarkable language understanding and generation capabilities on downstream tasks. However, when solving question answering (QA) tasks that require deeper reasoning and broader coverage, LLMs still face the constraints.
Owing to the high cost of re-training, the internal knowledge of LLMs cannot be updated in real time and thus remains limited in coverage, which results in failures to incorporate the latest information and increases the risk of hallucinations in downstream QA tasks~\cite{mallen2022not}. 

\footnotetext[1]{\url{https://github.com/ljy1228/SubQRAG}}
Retrieval-augmented generation (RAG)~\cite{lewis2020retrieval} addresses these challenges by retrieving knowledge from an external knowledge corpus. The standard RAG follows a single-step retrieval and generation framework. For simple QA tasks, this works well. However, this strategy is fragile for multi-hop QA because single-step retrieval can cause the retriever to narrowly focus on one dominant topic. Document-level knowledge often exhausts the limited retrieval window with evidence for a single hop, crowding out the distinct information needed to complete the full reasoning chain~\cite{zhuang2024efficientrag}. To advance beyond retrieving redundant document spans and to enable a more robust reasoning process, Graph RAG utilizes a knowledge graph (KG) structure. It compresses document-level knowledge into triples as $\langle entity, relation, entity \rangle$ to describe relationships between entities\cite{thompson2025inference}. This allows a guided traversal through the knowledge base, such as in GraphRAG~\cite{edge2024local}, facilitating the integration of multiple sources for complex QA.


Although graph RAG has advanced, it still faces limitations. A primary issue is its continued reliance on single-step retrieval, where the entire complex question is used to perform a single-step search for entry points into the graph. This approach fails to break down the reasoning process, making it difficult to dynamically adjust the retrieval focus for each logical hop~\cite{yao2022react}. This problem is further exacerbated by the static nature of the pre-constructed KG. Its structure is fixed and often incomplete, with partial coverage and coarse granularity weakening the reasoning support~\cite{wang2019tackling}. This inflexibility leads to a third major challenge: error accumulation. Because the initial entry nodes are determined by a single, holistic question, any imprecision can derail the entire reasoning chain. Subsequent pathfinding or node expansion inherits and amplifies these initial errors~\cite{cao2023rpa}, causing irrelevant exploration of the graph and ultimate failure~\cite{zhang2025credible}.

Motivated by these challenges, we propose SubQRAG, a framework designed to equip Graph RAG with two crucial capabilities: multi-step reasoning and real-time dynamic updating. To enable multi-step reasoning, SubQRAG first decomposes the original question into an ordered chain of sub-questions. It then addresses each sub-question sequentially, performing a focused retrieval of relevant triples at each step. To facilitate dynamic updating, if the pre-constructed graph lacks the necessary information, SubQRAG falls back to source documents. It retrieves relevant text, extracts new triples, and integrates them into the graph. Finally, all triples used in this adaptive reasoning process are aggregated into a coherent ``graph memory''. This memory provides a structured and traceable evidence path for the generator, mitigating the error accumulation inherent in undirected graph traversal.

Our contributions are summarized as follows:
\begin{itemize}
    \item \textbf{Decompose sub-questions for dynamic retrieval}. We present SubQRAG, which decomposes the original question into sub-questions and retrieves relevant triples, helping dynamically adjust the retrieval process and provide richer evidence.
    \item \textbf{Dynamic graph updating}. When the pre-constructed graph cannot answer the sub-questions, sub-questions drive KG updates by retrieving source documents.
   \item \textbf{Graph memory as structured guidance}. We integrate the triples supporting each sub-question into a subgraph and provide it to the generator. In this way, SubQRAG produces ``graph memory'' as reasoning trajectory, which reduces the limitations of the context window and improves QA performance. 

\end{itemize}

\begin{figure*}[!t]
  \centering
  \includegraphics[width=\textwidth, trim={70 310 0 360}, clip]{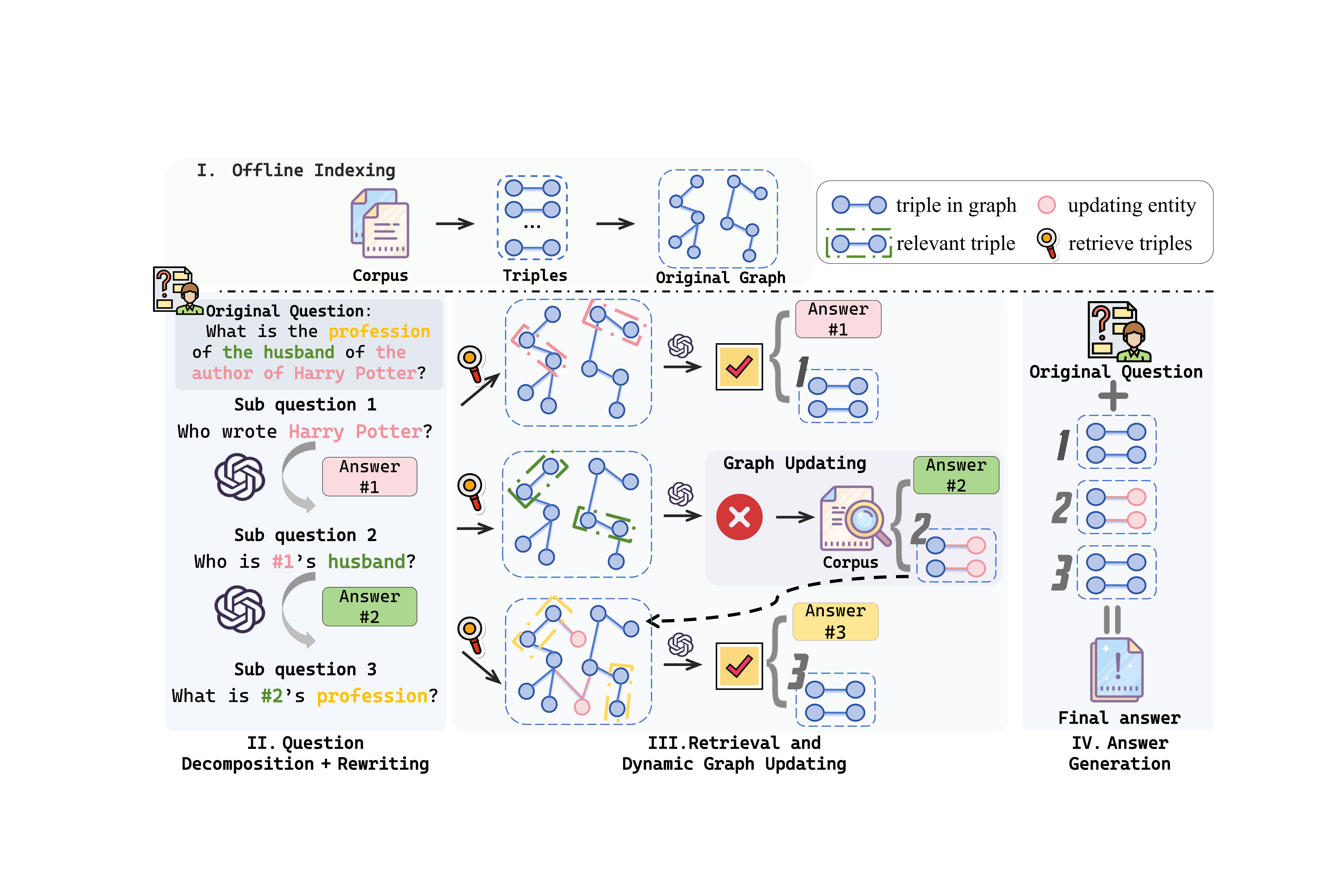}

  \caption{The SubQRAG framework consists of four stages: (I) Offline Indexing: the corpus is processed into triples to construct the initial KG. (II) Question Decomposition + Rewriting: the original question is decomposed into verifiable sub-questions. (III) Retrieval and Dynamic Graph Updating: each sub-question retrieves supporting triples from the graph. If unanswered, documents are consulted to extract new triples and update the graph. (IV) Answer Generation: supporting triples are assembled into graph memory to generate the final answer based on the original question.
}
  \label{fig:wide_results}
  \vspace{-2mm} 
\end{figure*}

\section{Method}
\label{sec:pagestyle}

As shown in Figure~\ref{fig:wide_results}, our proposed SubQRAG, a sub-question-driven dynamic Graph RAG framework, operates in four main stages. The process begins with (I) Offline Indexing, where LLMs pre-construct a KG by extracting structured knowledge triples from raw text. Next, during (II) Question Decomposition and Rewriting, a complex original question is broken down into a series of simpler, verifiable sub-questions; to maintain coherence, subsequent sub-questions are then rewritten to incorporate answers from preceding ones. According to simpler sub-questions, SubQRAG is able to perform (III) Retrieval and Dynamic Graph Updating, where for each sub-question, the framework attempts to retrieve supporting triples from the KG. Once the existing graph lacks the necessary information, SubQRAG will dynamically update the original graph by extracting new triples from the source documents. Finally, in the (IV) Answer Generation stage, all supporting triples gathered throughout the intermediate reasoning process are assembled into a graph-based memory, which is then provided to an LLM to synthesize a comprehensive final answer to the original question.

\subsection{Offline Indexing}

\noindent SubQRAG leverages an LLM to parse text documents, 
extract entities and their relations, and organize them into triples $\langle h, r, t \rangle$, which are then used to construct a corpus-level KG $\mathcal{G}$. The KG is formally defined in Eq.~~\ref{eq:triples}.
In this graph, nodes $h,t \in \mathcal{E}$ represent entities, while the relation edge $r \in \mathcal{R}$ captures the semantic link between the two entities. 
To avoid redundancy, SubQRAG checks whether a newly generated triple already exists, 
ensuring the conciseness of $\mathcal{G}$.

\begin{equation}
\mathcal{G} = \{ (h, r, t) \mid h, t \in \mathcal{E}, \; r \in \mathcal{R} \}
\label{eq:triples}
\end{equation}
\label{ssec:Offline}

\label{ssec:Online}
\subsection{Question Decomposition and Rewriting}

\subsubsection{Question Decomposition}
\label{sssec:decomposition}

In this stage, the goal of SubQRAG is to logically decompose the original multi-hop question $q$ into an ordered set of sub-questions $\{q_1, q_2, \dots, q_n\}$, where each $q_i$ may depend on the results of its predecessors $\{q_1, q_2, \dots, q_{i-1}\}$. This sequential dependency ensures that sub-questions are resolved in a logically consistent order.

\subsubsection{Question Rewriting}
\label{sssec:rewriti}

In the question rewriting stage, we explicitly incorporate the answers from previously resolved sub-questions into the current sub-question. As shown in Eq.~\ref{eq:rewrite}, for the $i$-th sub-question $q_i$, we leverage the answer set $A_{<i} = \{a_1, a_2, \dots, a_{i-1}\}$ obtained from prior steps to better guide the rewriting process. The function $\text{Rewrite}(\cdot)$ injects key entities or values from $A_{<i}$ into $q_i$, transforming it into a self-contained, unambiguous question $q_i'$ while preserving its original intent. Rewriting eliminates referential ambiguity and contextual gaps, improving retrieval precision and reasoning reliability.
\begin{equation}
q_i' = \text{Rewrite}(q_i, A_{<i})
\label{eq:rewrite}
\end{equation}

\subsection{Retrieval and Dynamic Graph Updating}
\label{sssec:RetrievalandDynamicGraphUpdating}

\noindent For each sub-question, SubQRAG employs an embedding model $f(\cdot)$  to compute semantic similarity between question $q_i$ and  triples in the pre-constructed KG. The top-k most relevant triples are selected shown in Eq.~\ref{eq:topk_main}. Based on these triples, generator obtains the answer and the used triples  $\mathcal{T}^{\mathrm{used}}_i $ to support subsequent sub-question answering:
\begin{equation}\label{eq:topk_main}
\mathcal{T}_{\mathrm{sub}} = 
\operatorname*{\text{Top-k}}_{r \in \mathcal{G}}
\cos\bigl(f(q_i),\, f((h, r, t))\bigr)
\end{equation}

Once the generator cannot retrieve sufficient information from the selected triples to respond, SubQRAG falls back to corpus-level retrieval. At this stage, we retrieve relevant documents and generate the answer based on their context. From these documents, new triples $\mathcal{T}^{\star}_{\mathrm{new}}$ are extracted, validated, and de-duplicated. The triples are then incrementally written back into the KG.
\begin{equation}\label{eq:newtriples}
\mathcal{T}^{\star}_{\mathrm{new}} = \text{Extract}( \mathcal{D} )
\end{equation}

This above process leverages sub-questions' answering to continually update the pre-constructed KG. Rather than the model’s parametric knowledge, each question is answered using only the retrieved triples.

\subsection{Answer Generation}

\label{sssec:answer_generation}

After sub-question reasoning, we collect all triples that are actually used in the intermediate reasoning process to construct the graph memory as shown in Eq.~\ref{eq:graphmemory}.
\begin{equation}
\label{eq:graphmemory}
\mathcal{T}^{\mathrm{used}} = \bigcup_{i=1}^{m} \mathcal{T}^{\mathrm{used}}_i 
\end{equation}

The final input to the generator consists of the original question $q$ together with the graph memory $\mathcal{T}^{\mathrm{used}}$.
Instead of document-level input, the generator leverages the context window more effectively and provides richer evidence. Meanwhile, this procedure yields graph memory in the form of traceable evidence paths, improving interpretability and reducing error accumulation.

\newcolumntype{L}{>{\raggedright\arraybackslash}X}
\newcolumntype{C}{>{\centering\arraybackslash}X}

\begin{table*}[t]
\centering
\begin{threeparttable}
\caption{EM and F1 scores on MuSiQue, 2Wiki, and HotpotQA multi-hop QA benchmarks. Zero-shot means evaluating the parametric knowledge of the backbone LLM without RAG. Retrieval-only baselines include embedding-based retrievers (GTR and all-MiniLM-L6-v2). Graph RAG baselines include RAPTOR, GraphRAG, LightRAG, and HippoRAG2. All graph RAG methods use all-MiniLM-L6-v2 as retriever and gpt-4o-mini as generator. This table highlights the best results in each column.}
\label{tab:gpt4omini_em_f1}

\begin{tabularx}{\textwidth}{
>{\centering\arraybackslash}m{2.5cm}
>{\centering\arraybackslash}m{3.5cm}
*{6}{>{\centering\arraybackslash}X}
}

\toprule
\multirow{2}{*}{\textbf{Baseline Type}} &
\multirow{2}{*}{\textbf{Method}} &
\multicolumn{2}{c}{\textbf{MuSiQue}} &
\multicolumn{2}{c}{\textbf{2Wiki}} &
\multicolumn{2}{c}{\textbf{HotpotQA}} \\
\cmidrule(lr){3-4}\cmidrule(lr){5-6}\cmidrule(lr){7-8}
& & \textbf{EM} & \textbf{F1} & \textbf{EM} & \textbf{F1} & \textbf{EM} & \textbf{F1} \\
\midrule

\multirow{1}{*}{Zero-shot} 
& None & 11.20 & 22.00 & 30.20 & 36.30 & 28.60 & 41.00 \\
\midrule

\multirow{2}{*}{Retrieval-only}
& GTR~\cite{ni-etal-2022-large} & 14.20 & 21.78 & 29.80 & 35.01 & 42.10 & 50.66 \\
& all-MiniLM-L6-v2~\cite{wang-etal-2021-minilmv2} & 24.10 & 36.38 & 29.80 & 35.01 & 42.10 & 50.67 \\
\midrule

\multirow{4}{*}{Graph RAG}
& RAPTOR~\cite{sarthi2024raptor} & 4.80  & 12.20 & 24.70 & 26.38 & 29.50 & 37.50 \\
& GraphRAG~\cite{edge2024local} & 4.69 & 13.30 & 35.37 & 40.13 & 28.97 & 39.33 \\
& LightRAG~\cite{guo2024lightrag} & 9.10  & 17.64 & 33.61 & 43.74 & 21.60  & 23.97 \\
& HippoRAG2~\cite{gutiérrez2025ragmemorynonparametriccontinual} 
   & 28.20 & \textbf{40.70} & 50.60 & 59.79 & 51.90 & \textbf{66.18} \\
\midrule

\multirow{1}{*}{Our proposed}
& \textbf{SubQRAG} & \textbf{29.70} & 38.14 & \textbf{61.90} & \textbf{64.30} & \textbf{56.00} & 64.30 \\
\bottomrule
\end{tabularx}
\end{threeparttable}
\end{table*}

\section{EXPERIMENTS}
\label{sec:typestyle}

\subsection{Datasets}

For evaluation, we adopt three widely used multi-hop question answering benchmarks: HotpotQA~\cite{yang-etal-2018-hotpotqa}, MuSiQue~\cite{trivedi2022musique} and 2WikiMultiHopQA~\cite{ho-etal-2020-constructing}. These datasets cover different reasoning scenarios, including cross-document reasoning, complex contextual understanding and reasoning. This provides a comprehensive evaluation of SubQRAG on multi-hop QA tasks.
To ensure both reproducibility and fairness, we strictly follow the experimental protocol of HippoRAG2~\cite{gutiérrez2025ragmemorynonparametriccontinual}, using the same retrieval corpus and sampling 1,000 questions from each validation set as test questions. Under this unified setup, the performance of SubQRAG can be fairly compared against existing methods.
\subsection{Baselines}

We compare against three categories of baselines under a unified experimental setting. Zero-shot reports the performance of the backbone language model without RAG. RAG evaluates strong embedding-based retrievers, including GTR~\cite{ni-etal-2022-large} and all-MiniLM-L6-v2~\cite{wang-etal-2021-minilmv2} , serving as natural baselines. Graph RAG covers structure-enhanced approaches: RAPTOR~\cite{sarthi2024raptor} builds a semantic hierarchy, GraphRAG~\cite{edge2024local} and LightRAG~\cite{guo2024lightrag} use KGs for concept-level summaries, and HippoRAG2~\cite{gutiérrez2025ragmemorynonparametriccontinual} further augments a schema-less KG with passage nodes, context edges, and dense–sparse retrieval, providing a strong multi-hop QA baseline.


\begin{table}[h]
\centering
\caption{Ablation performance on HotpotQA. 
w/o Decomposition removes the sub-question decomposition module. 
w/o Rewriting disables the question rewriting step. 
And w/o Update omits the dynamic updating graph.}
\begin{threeparttable} 

\label{tab:ablation}
\begin{tabular}{lcc}
\toprule
\textbf{Method} & \textbf{EM} & \textbf{F1} \\
\midrule
SubQRAG & \textbf{56.0} & \textbf{64.3} \\
w/o Decomposition & 50.5 (-5.5) & 59.6 (-4.7) \\
w/o Rewriting & 49.5 (-6.5) & 50.2 (-14.1) \\
w/o Update & 54.5 (-1.5) & 63.7 (-0.6) \\
\bottomrule
\end{tabular}
\end{threeparttable}
\end{table}

\subsection{Metrics}

We report two standard QA metrics. \textbf{Exact Match (EM)} measures whether the model’s answer exactly matches the gold answer after simple normalization, serving as a strict accuracy metric. \textbf{F1} score computes the harmonic mean of precision and recall between the predicted and gold answers. 

\subsection{Implementation Details}

For SubQRAG, we employ gpt-4o-mini as the backbone model across all key stages, including entity recognition, knowledge extraction, sub-question decomposition, triple construction, question rewriting and question answering. For the embedding component, we adopt all-MiniLM-L6-v2 as the retriever to encode and match triples and at each retrieval step the top-5 triples are selected to support answer generation. To ensure a fair comparison, all baselines, except zero-shot and GTR, are equipped with the same retriever and generator. All experiments for SubQRAG are conducted on a single NVIDIA RTX 3090 GPU.

\subsection{Results}
As shown in Table~\ref{tab:gpt4omini_em_f1}, we evaluate various retrievers across three benchmarks using gpt-4o-mini as the QA generator, comparing zero-shot, retrieval-only, graph-based RAG methods and our proposed SubQRAG. For zero-shot, although employing gpt-4o-mini, it still fails to achieve strong results. For standard RAG, the results on MuSiQue and HotpotQA show partial improvements owing to retrieving knowledge from an external knowledge base. Graph-based approaches that capture relationships among documents further improve performance, especially HippoRAG2. However, these methods still lack the ability to dynamically update the graph, and perform deeper multi-step reasoning. For our improved SubQRAG, it achieves consistent improvements on EM scores across three benchmarks, with gains of 5.3\% on MuSiQue, 22.3\% on 2Wiki, and 7.9\% on HotpotQA. 


\subsection{Ablation Study}
We design ablation experiments for the proposed SubQRAG on HotpotQA. Table~\ref{tab:ablation} shows that removing decomposition, rewriting, and graph updating all lead to performance drops. Decomposition and rewriting are the key factors, while graph updating has a relatively modest effect since the KG is infrequently updated, but it still provides gains, confirming the effectiveness of sub-question driven updates.
 
\section{Conclusion}

In conclusion, we propose SubQRAG, a sub-question driven graph RAG framework which consists of four stages. Based on pre-construct KG, it first decomposes complex questions into verifiable sub-questions, then dynamically updates the KG when necessary for fine-grained retrieval and finally provides structured ``graph memory'' to the generator. Through leveraging multi-step reasoning and real-time graph updating, SubQRAG addresses incompleteness and error accumulation in the static graph RAG. Experiments on multi-hop QA benchmarks show that SubQRAG consistently surpasses existing graph RAG approaches, demonstrating its effectiveness in multi-hop QA tasks and achieving balance between the reasoning depth and knowledge base breadth.



\vfill\pagebreak

\bibliographystyle{IEEEbib}
\bibliography{strings,refs}

\end{document}